\documentclass{article}

\usepackage[preprint]{neurips_2026}

\usepackage[utf8]{inputenc} 
\usepackage[T1]{fontenc}    
\usepackage{hyperref}       
\usepackage{url}            
\usepackage{booktabs}       
\usepackage{amsfonts}       
\usepackage{nicefrac}       
\usepackage{microtype}      
\usepackage{xcolor}         
\usepackage{natbib}
\usepackage{graphicx}
\usepackage{float}
\usepackage{enumitem}
\usepackage{wrapfig}
\newcommand{\modelicon}[1]{\raisebox{-0.15em}{\includegraphics[height=1.05em]{#1}}}
\title{Can Agents Price a Reaction? \\ Evaluating LLMs on Chemical Cost Reasoning}

\author{%
  \textbf{Yuyang Wu}\textsuperscript{1}\thanks{Equal contribution.}\hspace{1.2em}
  \textbf{Yue Huang}\textsuperscript{2}\footnotemark[1]\hspace{1.2em}
  \textbf{Shuaike Shen}\textsuperscript{1}\hspace{1.2em}
  \textbf{Xujian Wang}\textsuperscript{1} \\
  \textbf{Shuhao Zhang}\textsuperscript{1}\hspace{1.2em}
  \textbf{Qiyao Xue}\textsuperscript{3}\hspace{1.2em}
  \textbf{Weichen Liu}\textsuperscript{4}\hspace{1.2em}
  \textbf{Runtian Gao}\textsuperscript{1} \\
  \textbf{Jian Ma}\textsuperscript{1}\thanks{Corresponding authors.}\hspace{1.2em}
  \textbf{Xiangliang Zhang}\textsuperscript{2}\footnotemark[2]\hspace{1.2em}
  \textbf{Olexandr Isayev}\textsuperscript{1}\footnotemark[2] \\[3pt]
  \textsuperscript{1}Carnegie Mellon University \hspace{1em}
  \textsuperscript{2}University of Notre Dame \\
  \textsuperscript{3}University of North Carolina, Chapel Hill \hspace{1em}
  \textsuperscript{4}University of Pittsburgh
}

\begin{document}

\maketitle

\begin{abstract}

Large Language Models (LLMs) have become increasingly capable as tool-using agents, with benchmarks spanning diverse general agentic tasks. Yet rigorous evaluation of scientific tool use remains limited. In chemistry, recent agents can plan syntheses and invoke domain-specific tools, but evaluations often rely on curated demonstrations, expert assessment, or LLM-as-judge scoring rather than exact, judge-free ground truth. We address this gap with chemical procurement cost estimation, a practical task in which an agent must ground chemical identities, retrieve supplier quotes, select valid purchasable packs, normalize quantities, and compute cost from a reaction description. We introduce \textsc{ChemCost}, a benchmark of 1,427 evaluable reactions grounded to a frozen pricing snapshot covering 2,261 chemicals and 230,775 supplier quotes,  supporting scalar scoring and stage-level diagnosis of grounding, retrieval, procurement, and arithmetic failures. To evaluate robustness, we further construct controlled noise-injected views that perturb chemical aliases, quantity expressions, missing fields, and input formatting. Experiments with frontier, open-weight, and chemistry-specialized LLM agents show that tool access is necessary but insufficient for solving the task. The strongest agents reach only 50.6\% accuracy within 25\% relative error on clean inputs and degrade substantially with realistic noise. Stage-level analysis further shows that failures arise from brittle parsing, ineffective evidence integration, invalid pack selection, and non-convergent tool use.
\end{abstract}

\section{Introduction}







Large language models (LLMs) have demonstrated strong capabilities as tool-using agents. When integrated with agentic frameworks, they can plan over intermediate states, invoke external tools, revise actions from feedback, and complete multi-step tasks through iterative interaction~\cite{yao2022react}. Prior work has evaluated these capabilities through benchmarks such as SWE-bench, WebArena, and ToolLLM~\cite{jimenez2023swe, zhou2023webarena,qin2023toolllm}, which test agents in software engineering, web interaction, and tool invocation environments. While these benchmarks have been instrumental in measuring general agentic competence, they primarily focus on broadly applicable tool orchestration and typically require limited domain-specific reasoning beyond the information exposed by the prompt or environment.

Beyond general-purpose settings, LLMs have also shown growing promise in scientific domains, particularly chemistry. Recent systems such as ChemCrow~\cite{bran2023chemcrow} and Coscientist~\cite{boiko2023autonomous} demonstrate that language models can support chemical synthesis planning, invoke domain-specific tools and query external resources to assist chemical experiment. However, many scientific-agent evaluations remain centered on curated demonstrations, expert assessment, or high-level task success, leaving comparatively less attention to real-world industrial workflows that require agents to interact with structured databases, satisfy domain constraints, and perform quantitatively grounded decision-making. Chemical procurement cost estimation is a representative example of such a workflow. In chemical industry and laboratory planning, estimating the cost of producing a target compound is essential for route evaluation, budget planning, supplier selection, and feasibility assessment. This task requires an agent to connect chemical knowledge with practical procurement information, rather than only producing a textual answer from parametric knowledge.

\begin{figure*}[t]
  \centering
  \includegraphics[width=1.0\textwidth, height=0.55\textwidth]{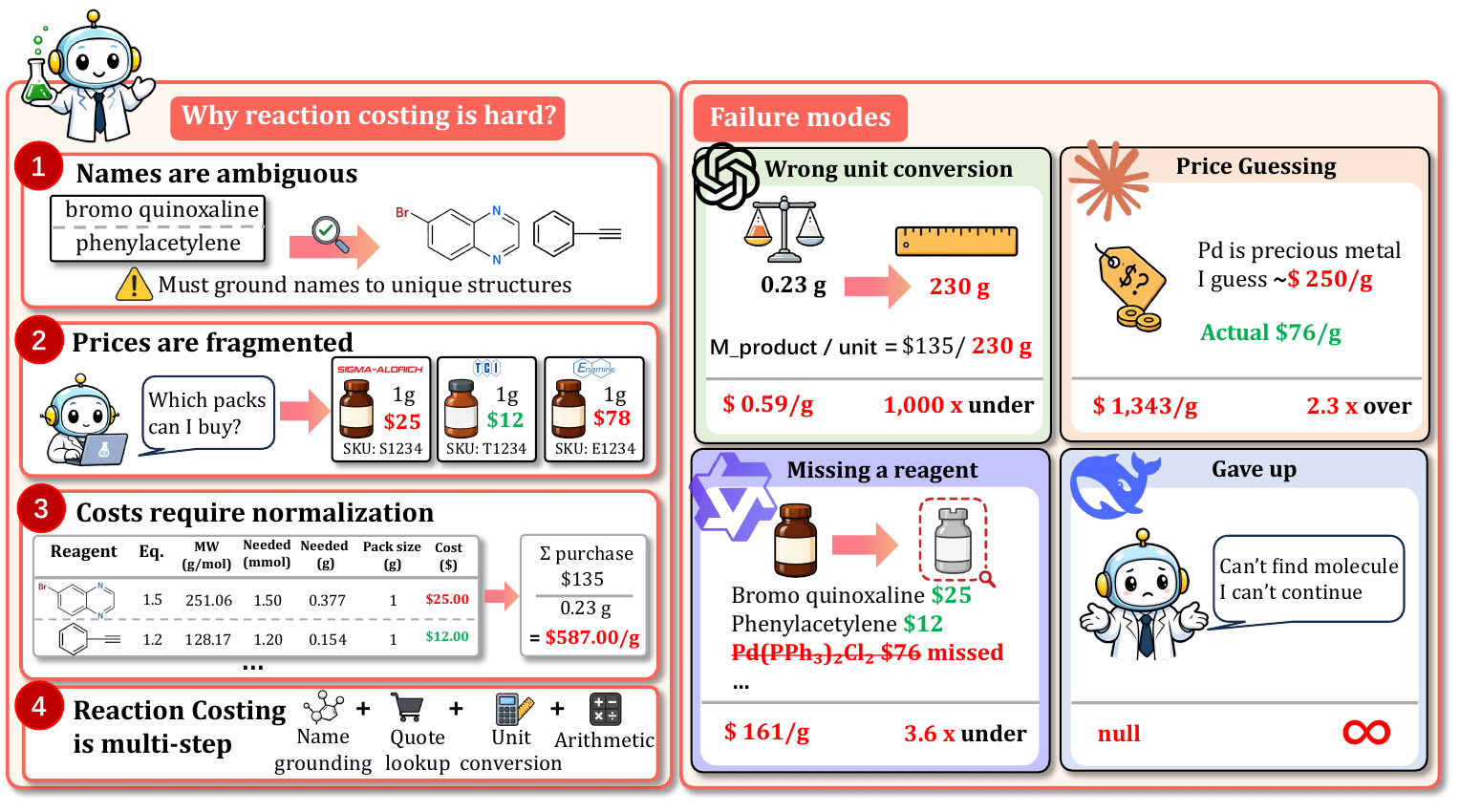}
  \vspace{-0.33in}
  \caption{Overview of the procurement cost estimation task and representative failure modes.}
  \label{fig:figure1}
  \vspace{-0.27in}
    
\end{figure*}

However, chemical procurement cost estimation is challenging for LLM agents as it requires a sequence of interdependent, domain-constrained operations, illustrated in Figure~\ref{fig:figure1}. Given a target compound and a reaction specification, an agent must resolve ambiguous chemical names, retrieve supplier quotes from a pricing database, select valid purchasable packs under purity and quantity constraints, convert reaction ratios into required masses, and aggregate component costs under a specified yield assumption. Errors can arise at any stage, the agent may map a compound to the wrong identifier, retrieve an incorrect supplier entry, choose an invalid pack, mishandle units or stoichiometric ratios, or perform incorrect arithmetic aggregation. At the same time, this structure makes the task highly suitable for diagnostic evaluation, since each intermediate step produces an output that can be checked automatically against ground truth. Chemical procurement cost estimation therefore provides a realistic and stage-wise verifiable testbed for evaluating scientific agents under practical tool-use and domain-reasoning constraints.

To address this gap, we introduce \textsc{ChemCost}, a benchmark for evaluating LLM agents on chemical procurement cost estimation. Each instance presents a reaction description while withholding supplier prices and standardized molecular identifiers; the agent must return the procurement cost per gram of product normalized from a fixed 1 mmol limiting-reagent scale. \textsc{ChemCost} comprises 1,427 curated reaction records drawn from five complementary sources: structured reaction databases, named-reaction textbooks, \textit{Organic Syntheses} procedures, retrosynthesis route collections, and hand-crafted examples. Ground truth is computed deterministically from a frozen snapshot of 230,775 pack-level supplier quotes spanning 2,261 chemicals under a fixed procurement rule. Solving an instance requires grounding chemical names, identifying the limiting reagent, retrieving and selecting qualifying supplier quotes, and aggregating stoichiometric procurement costs. Agents interact via tools for name resolution, quote retrieval, molecular-weight computation, and arithmetic. Evaluation is fully judge-free, with stage-level intermediates verified automatically against the frozen database. To stress-test robustness under realistic input variation, we introduce a four-layer noise-injection pipeline that perturbs chemical names, reaction quantities, side information, and surface format while preserving ground truth.

Beyond aggregate accuracy, \textsc{ChemCost} supports stage-level analysis of how scientific tool use succeeds and fails. Logged trajectories reveal that tool engagement is necessary but insufficient: active agents can fail after many calls via non-convergent supplier retrieval or premature termination. Noise injection shows format perturbations are the strongest abstention trigger, while null predictions arise through distinct mechanisms across models, including budget exhaustion, non-completion, and refusal. Structurally, route depth is the dominant difficulty factor, with component count and product molecular weight contributing model-dependent effects. Together, these findings frame procurement reasoning as a structured tool-use challenge requiring reliable grounding, evidence integration, and quantitative state tracking under noisy inputs.

Our contributions can be summarized as follows:
\begin{itemize}[leftmargin=*]
\item We formalize chemical procurement cost estimation as a practical and structurally challenging scientific tool-use task, where the agent must recover a computable procurement state from reaction text by grounding ambiguous chemicals, retrieving fragmented supplier quotes, selecting valid packs, normalizing quantities, and performing exact cost aggregation.

\item We introduce \textsc{ChemCost}, a benchmark of 1,427 evaluable reactions grounded to a frozen supplier database covering 2,261 chemicals and 230,775 pack-level quotes. Each label is computed deterministically from procurement rules, enabling exact scoring without human or LLM judges.

\item We design a controlled evaluation protocol with clean structured inputs, noise-injected views, and stage-level diagnostics for component recognition, tool trajectories, abstention, and cost accuracy. The noise suite perturbs chemical aliases, quantity expressions, missing fields, and input formats while preserving the same underlying costing target.

\item We provide a stage-level diagnostic analysis of scientific tool use across frontier, open-weight, and chemistry-specialized LLM agents. We find that procurement reasoning depends on how agents integrate retrieved evidence, not merely whether they call tools. Format noise emerges as the strongest abstention trigger, while route depth is the dominant structural driver of task difficulty.

\end{itemize}

\section{Related Work}
\label{related_work}


\textbf{Tool-Use Agent Benchmarks.} A growing body of work evaluates the ability of LLMs to function as tool-using agents that interact with external environments.  Existing benchmarks instantiate this problem in several realword application settings. Software-engineering benchmarks, such as CodeAgentBench~\cite{zhang2024codeagent}, evaluate agents on code-generation tasks that require repository retrieval, code-symbol navigation, and testing. Web and computer-use environments, such as WebArena~\cite{zhou2023webarena}, emphasize multi-step navigation and tool use under changing page states. Structured tool-calling benchmarks, including ToolLLM, T-Eval, METATOOL, and AppWorld~\cite{qin2023toolllm, chen2024t, huang2023metatool, trivedi2024appworld}, focus more directly on API-level behavior, such as tool selection, argument construction, tool composition, and execution feedback. Together, these benchmarks have substantially advanced the evaluation of planning, tool orchestration, and environment grounding. These benchmarks primarily evaluate general-purpose interaction skills, where tool-relevant information is often exposed in the environment or task instruction. Scientific domains pose a different challenge, as effective tool use usually requires domain knowledge, entity resolution, and interpretation of domain specialized inputs.



\textbf{Scientific Reasoning Benchmarks and Chemistry Agents.}
Moving beyond general-purpose tool-use settings, scientific domains require agents to combine tool interaction with domain-specific reasoning. Scientific LLM benchmarks, including SciBench, ChemLLMBench, MolErr2Fix, and ChemCoTBench~\cite{wang2023scibench, guo2023can, wu2025molerr2fix, li2025beyond}, evaluate scientific knowledge, quantitative reasoning, molecular understanding, and chemistry-specific problem solving through multiple-choice, free-form QA, and error-correction tasks. These benchmarks test scientific reasoning over textual inputs, but generally do not require long-horizon tool interaction~\cite{song2025evaluating}. In parallel, chemistry agents such as ChemCrow, Coscientist, LLM-RDF, and CheMatAgent~\cite{m2024augmenting, boiko2023autonomous, ruan2024automatic, wu2025chematagent} show that LLMs can plan syntheses, invoke chemistry tools, query structured resources, and interface with laboratory automation. These systems demonstrate the promise of tool-augmented scientific agents, but lack a controlled benchmark for domain-grounded tool selection, multi-step execution, and judge-free outcome verification. Our benchmark addresses this gap through chemical procurement tasks, requiring agents to resolve chemical entities, retrieve supplier information, select valid packs, and aggregate costs under rules.



\textbf{Chemical Synthesis Planning with Cost Estimation.} Cost has long been recognized as a practical constraint in synthesis planning and pharmaceutical development. Prior work like Chematica, ASKCOS, and RouteScore~\cite{klucznik2018efficient, tu2025askcos, seifrid2022routescore} has incorporated cost into retrosynthesis and route evaluation in several ways, including reagent buyability heuristics, fixed catalog prices during search, constrained route optimization, and structured route-scoring frameworks. However, in these settings cost is typically treated as structured input to the planner rather than as a quantity that must be recovered by an agent. Pack identities are assumed resolved, prices are assumed retrieved, and limiting reagents are inferred from structured reaction representations instead of free text. Different from prior work, \textsc{ChemCost} evaluates whether a general-purpose LLM agent can recover procurement cost from textual reaction descriptions by resolving chemical identities, retrieving supplier quotes, selecting qualifying packs, and performing cost aggregation.

\section{ChemCost Benchmark}
\label{headings}
\begin{figure*}[t]
  \centering
  \includegraphics[width=1.0\textwidth]{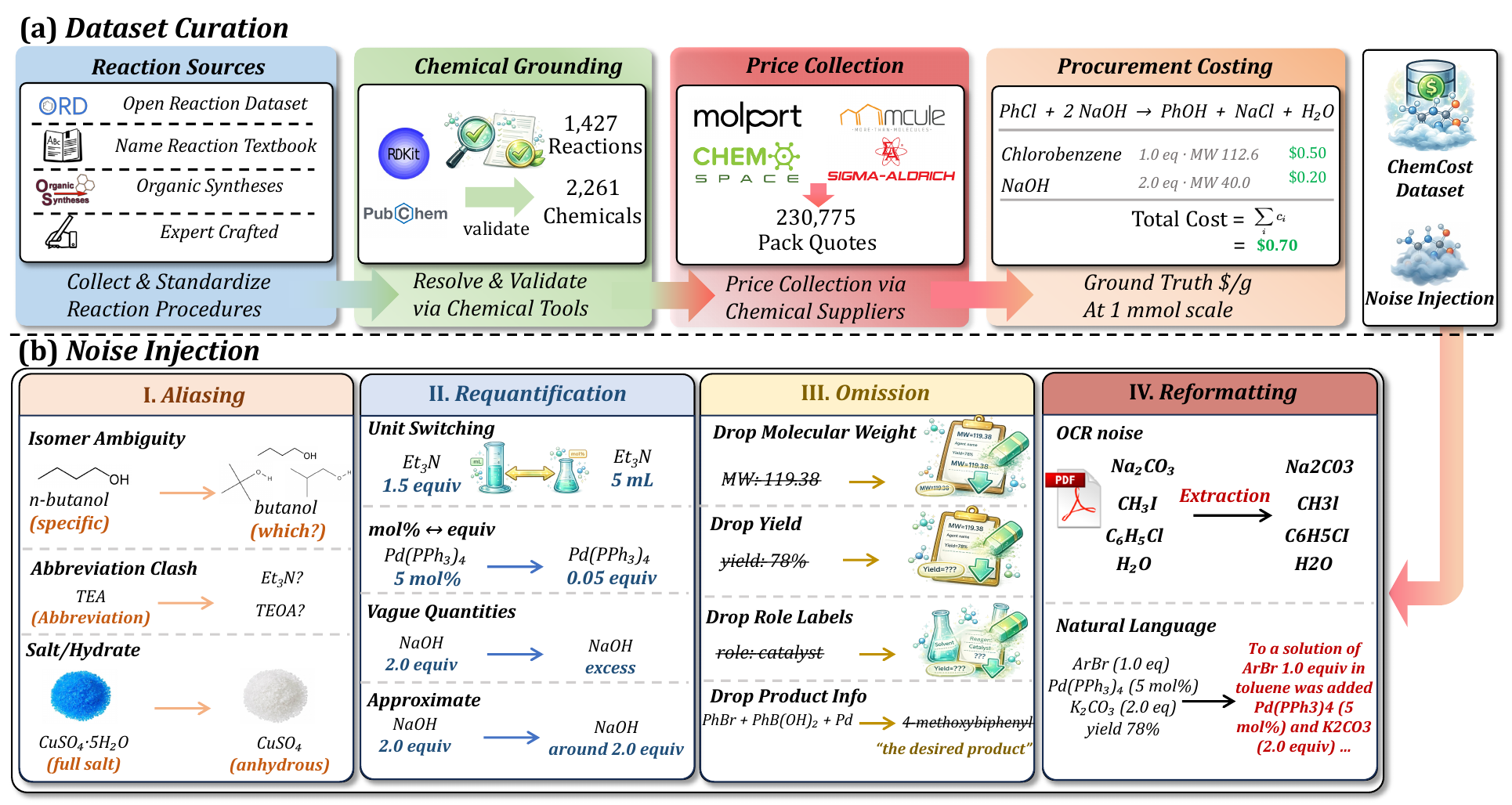}
  \vspace{-0.25in}
  \caption{\textbf{Data curation and noise-injection pipeline.} 
(a) Construction of \textsc{ChemCost} from reaction sources to deterministic cost labels. 
(b) Noise injection over four domains.}
  \label{fig:figure2}
  \vspace{-0.2in}
\end{figure*}

In this section, we introduce \textsc{ChemCost}, the first benchmark on evaluating LLM agents on multi-step chemical procurement cost estimation. As shown in Figure~\ref{fig:figure2}, the benchmark comprises curated reaction instances with deterministic labels derived from frozen supplier data and explicit procurement rules, along with a controlled noise-injection pipeline for agent robustness evaluation.

\subsection{Benchmark Construction}
\label{Benchmark Construction}
\textbf{Reaction Sources Collection.} \textsc{ChemCost} draws 1,427 reactions from five complementary source families. Structured records come from the Open Reaction Database, named-reaction entries from organic chemistry textbooks, and experimentally vetted procedures from \textit{Organic Synthesis}\cite{kearnes2021open, li2004name, smith2024organic}. Route-level collections are taken from PaRoutes and ChemPU~\cite{genheden2022paroutes, rohrbach2022digitization}, and a set of hand-crafted calibration examples. Each source is processed under a source-specific extraction rule as shown in Table~\ref{tab:sources} and passes through yield, structure, and cost-component filters.


\textbf{Reaction Extraction and Normalization.} All sources are first converted into a unified reaction schema containing a reaction identifier, product specification, yield, and component list. Component roles are normalized into a fixed inventory: \textit{reactant}, \textit{catalyst}, \textit{base}, \textit{reagent}, and \textit{solvent}. Quantities are expressed as equivalents relative to the limiting reagent, with catalytic loadings in mol\% converted as $e_i=\mathrm{mol\%}/100$. After this normalization, the limiting reagent has equivalent 1.0, so a 1 mmol limiting-reagent scale assigns component \(i\) an amount of \(e_i\) mmol. Solvents are excluded from cost computation but retained for component-recognition evaluation. We filter records with missing products, invalid yields, or no costed non-solvent components before computing ground-truth costs.

\textbf{Chemical Grounding.} After extraction, each chemical mention is grounded to a validated molecular identity. We resolve component and product names to canonical SMILES using PubChem~\cite{kim2023pubchem, weininger1988smiles}, and validate the resulting structures with RDKit~\cite{landrum2013rdkit}. Molecular weights are computed from the validated structures and stored in each benchmark record. If a source already provides SMILES, we still canonicalize and validate it to ensure consistent representation. As a result, downstream steps, including required-mass computation, supplier-quote retrieval, and component matching, are anchored to molecular identities rather than surface-form names.

\textbf{Price Collection.} For each grounded chemical, we collect commercial supplier quotes from marketplace and vendor-aggregation sources, including MolPort, ChemSpace, and Sigma-Aldrich~\cite{cihan2025chemprice}. Quotes are stored in a frozen pack-level database, with each entry recording the chemical identifier, supplier, pack size, price, purity, and snapshot metadata. To ensure reproducibility, all prices are fixed at a single snapshot. We apply basic validity filters to remove unusable entries, including missing prices, non-positive pack sizes, and quotes below the minimum purity threshold of 95\%, matching the standard research-grade specification of the collected supplier sources.

\textbf{Procurement Cost Calculation.}
For each curated reaction, \textsc{ChemCost} computes a deterministic fixed-scale procurement cost normalized per gram of product. We first normalize all equivalents relative to the limiting reagent, so that the limiting reagent has equivalent \(1.0\), and then fix the reaction scale to \(1\) mmol of the limiting reagent. For each non-solvent component \(i\), with normalized equivalent \(e_i\) and molecular weight \(M_i\), the required mass at this scale is \(m_i = e_i M_i \times 10^{-3}\) g. Its purchase cost \(q_i\) is determined by a fixed pack-selection rule: choose the smallest valid pack whose quantity covers \(m_i\); if no single pack covers \(m_i\), purchase the minimum number of largest valid packs required to cover \(m_i\). Given product molecular weight \(M_P\) and yield \(y\), the final label is \(c = \sum_{i:\rho_i \ne \mathrm{solvent}} q_i /(M_P \times 10^{-3} \times y/100)\). This label should be interpreted as the procurement cost of the \(1\) mmol-scale reaction normalized by the grams of product obtained at that scale, not as the exact purchasing cost required to scale the reaction to produce exactly \(1\) gram of product.

\vspace{-0.1in}
\subsection{Noise Injection}
\label{noise_injection}
\textbf{Stage 1: Chemical Name Noise.} Chemical names are highly redundant: the same molecule may appear as an abbreviation, common name, systematic IUPAC name, or formula, and these conventions are often mixed within a procedure. Positional or stereochemical prefixes may also be omitted when the intended isomer is obvious to human readers, creating ambiguity for agents. For example, \texttt{TEA} may refer to triethylamine or triethanolamine, which differ in molecular weight and procurement cost. This stage tests whether agents can resolve noisy surface forms to the correct molecular identity.

\textbf{Stage 2: Quantity Noise.} Stoichiometric information is often reported inconsistently across chemistry procedures. Catalyst loadings may be given as equivalents or molar percent, liquid reagents are frequently specified by volume rather than amount, and procedures may include approximate or qualitative descriptions such as \emph{ca. 1.3 equiv}, \emph{excess}, or \emph{a few drops}. These variations directly affect cost estimation, since the procurement model requires normalized component masses. Introducing such quantity noise can evaluate whether agents can convert heterogeneous quantity expressions into usable mass estimates, or abstain when the required information is genuinely unrecoverable.

\textbf{Stage 3: Missing-Information Noise.}
Real experimental procedures often omit structured fields that are useful for procurement-cost computation, such as molecular weights, yields, and component roles. These omissions do not change the underlying reaction or its oracle cost label, but usually force the agent to recover the missing procurement state from the remaining reaction description and available tools. Molecular weights can be recomputed from grounded structures, yields can be recovered from procedural or product information when available, and component roles must be inferred from chemical context.

\textbf{Stage 4: Format Noise.} Chemical information is rarely encountered in a clean structured schema. In practice, reaction data appears as free-text experimental prose, scanned documents with OCR artifacts, degraded Unicode text, or mixed narrative-and-table layouts. These presentation changes can break exact-match retrieval even when the underlying chemistry is unchanged, for example when CH$_3$I becomes \emph{CH3l} or Na$_2$CO$_3$ becomes \emph{Na2C03}. We apply such format transformations to test whether the agent's procurement workflow remains robust under realistic presentation formats rather than depending on a convenient benchmark schema.

\subsection{Evaluation Pipeline} 
ChemCost evaluates agents under a fixed reaction-to-cost protocol. For each scored instance, the evaluator removes information that would trivialize the task, including supplier prices, canonical component identifiers, procurement labels, cost tiers, and price statistics. The agent receives an agent-facing reaction view and must return a scalar prediction \(\hat{c}\), the procurement cost of product. The evaluator parses the final response for a valid positive numerical prediction. Explicit refusals, non-numerical responses, runtime failures, and step-budget exhaustion are recorded as null predictions. Null predictions remain in the denominator of tolerance-based accuracy metrics and abstention rate is reported separately. Cost tolerance accuracy is defined as $\mathrm{CTA}@k = \frac{1}{N}\sum_{j=1}^{N} \mathbf{1}[\,|\hat{c}_j-c_j|/c_j \leq k/100\,]$. In addition to CTA metrics, the evaluator logs complete tool trajectories to support stage-level failure attribution and reports diagnostic metrics including component precision, component recall, tool efficiency, token efficiency, and price-optimization score when recoverable.








\section{Experiments}
\subsection{Experimental Setup}
\label{Experimental Setup}
We evaluate all agents on the ChemCost benchmark under a unified reaction-to-cost protocol. For each reaction, the agent receives the same agent-facing view consisting of component names, component roles, stoichiometric equivalents, molecular weights, reported yield, and product identity. The agent must return a single scalar cost prediction. Supplier prices and canonical chemical identifiers are withheld from the input so that correct prediction requires explicit chemical resolution and multi-step procurement reasoning.

\textbf{Agents.}
We evaluated all models as tool-augmented ReAct agents. Each agent is equipped with four deterministic tools that implement the ChemCost procurement workflow, including chemical name resolution, frozen supplier-quote retrieval, molecular-weight computation, and arithmetic evaluation. The agent interacts with these tools through a ReAct reasoning-action loop with a maximum budget of 40 steps and must return a single scalar estimate of procurement cost. We evaluate frontier models, open-weight models, and chemistry-specialized models, including Claude Sonnet 4.6, GPT-5, GPT-5.4-mini, Gemini2.5-Flash, DeepSeek V4 Pro, Kimi K2.5, Qwen3.5-Plus, Qwen3-235B-A22B, Qwen3-14B, LlaSMol-7B, ChemDFM-v2.0-14B, and ChemLLM-20B~\cite{achiam2023gpt, team2023gemini, liu2024deepseek, team2025kimi, bai2023qwen, yu2024llasmol, zhao2024chemdfm, zhang2024chemllm}.

\textbf{Noise Protocol.} We evaluate agents under one clean view and five perturbed views. The \textsc{Clean} setting uses the structured reaction input directly. The four single-noise settings apply one medium-level perturbation stage: chemical-name aliasing (\textsc{+Name}), quantity re-expression (\textsc{+Qty}), missing-field perturbation (\textsc{+Miss}), and format perturbation (\textsc{+Fmt}). The \textsc{All Noise} setting applies all four stages jointly. Ground-truth labels are unchanged across views.

\textbf{Metrics.} The primary evaluation metric is cost tolerance accuracy, CTA@$k$, defined as the fraction of reaction whose predicted procurement cost falls within $k\%$ of the deterministic ground truth. We report CTA@10 and CTA@25 in the clean setting, and CTA@10 across noise settings. In addition, we report component precision and component recall by matching predicted component lists against the ground-truth component set using the benchmark's synonym and abbreviation alias index.

\subsection{Experimental Results}
\label{Experimental Results}
Table~\ref{tab:main} reports the main results for tool-augmented ReAct agents. On clean structured inputs, the strongest models achieve moderate cost accuracy. Qwen3.5-Plus obtains the highest CTA@25 at 50.6\%, but still remains below the human reference. Performance drops under the stricter CTA@10 metric, where DeepSeek V4 Pro achieves the best score at 37.0\%. These results indicate that agents can often recover approximate procurement costs, but reliable numerical accuracy remains limited.


Performance varies across model families. Large frontier and open-weight models dominate the clean setting, with Qwen3.5-Plus, DeepSeek V4 Pro, GPT-5, and Kimi K2.5 exceeding 37\% CTA@25. Smaller or base models are weaker. GPT-5.4 Mini reaches 11.5\% CTA@25, Qwen3-14B reaches 9.9\%, and Qwen3-235B-A22B reaches 6.2\%. Chemistry-specialized models do not consistently improve procurement-cost accuracy. Although ChemDFM-v2.0-14B and ChemLLM-20B achieve high component precision and recall, their CTA scores remain near zero in the clean setting. This suggests that chemistry-specific fine-tuning improves chemical recognition, but does not transfer to agentic procurement reasoning, where models must follow instructions, use tools, track intermediate state, and perform quantitative aggregation.

Across noise-injected views, performance decreases relative to the clean setting. For the strongest agents, chemical-name and quantity perturbations are less damaging than format perturbations. Qwen3.5-Plus maintains 30.9\% CTA@10 under +Name and 34.6\% under +Qty, but drops to 18.5\% under +Fmt and 19.8\% under All Noise. DeepSeek V4 Pro shows a similar pattern, achieving 38.3\% under +Name and 37.0\% under +Qty, but only 22.2\% under +Fmt and 19.8\% under All Noise. These results show that robustness to presentation formats remains a major bottleneck.

\begin{table*}[t]
\centering
\footnotesize
\caption{\textbf{\textsc{ChemCost} main results.} Tool-augmented ReAct agents are evaluated on clean inputs and noise-injected views. Pre. and Rec. denote precision and recall; C.10 and C.25 denote CTA@10 and CTA@25. The human row is a reference and is excluded from automated-agent boldface comparisons. All Noise applies all four perturbation types jointly. The best results are in \textbf{bold}.}
\label{tab:main}
\setlength{\tabcolsep}{1.8pt}
\renewcommand{\arraystretch}{0.92}
\resizebox{\textwidth}{!}{%
\begin{tabular}{@{}l cccc @{\hspace{7pt}} ccc @{\hspace{7pt}} ccc @{\hspace{7pt}} ccc @{\hspace{7pt}} ccc @{\hspace{7pt}} ccc@{}}
\toprule
 & \multicolumn{4}{c}{Clean} & \multicolumn{3}{c}{+Name} & \multicolumn{3}{c}{+Qty} & \multicolumn{3}{c}{+Miss} & \multicolumn{3}{c}{+Fmt} & \multicolumn{3}{c}{All Noise} \\
\cmidrule(lr){2-5} \cmidrule(lr){6-8} \cmidrule(lr){9-11} \cmidrule(lr){12-14} \cmidrule(lr){15-17} \cmidrule(lr){18-20}
Models & Pre. & Rec. & C.10 & C.25 & Pre. & Rec. & C.10 & Pre. & Rec. & C.10 & Pre. & Rec. & C.10 & Pre. & Rec. & C.10 & Pre. & Rec. & C.10 \\
\midrule
Human                  & --- & --- & 57.0 & 66.9 & --- & --- & 55.3 & --- & --- & 55.0 & --- & --- & 43.3 & --- & --- & 46.7 & --- & --- & 36.6 \\
\midrule
\multicolumn{20}{@{}l}{\textit{Closed-source}} \\
\modelicon{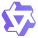}Qwen3.5-Plus           & 100.0 & 54.3 & 34.6 & \textbf{50.6} & 98.2 & 51.8 & 30.9 & 99.6 & 57.9 & 34.6 & 99.6 & 54.1 & 32.1 & 69.9 & 20.4 & 18.5 & 72.2 & 19.7 & \textbf{19.8} \\
\modelicon{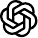}GPT-5                  & 93.5 & 32.8 & 24.7 & 38.3 & 88.4 & 31.2 & 23.5 & 94.2 & 29.6 & 23.5 & 92.2 & 29.6 & 22.2 & 59.6 &  7.0 & 13.6 & 65.8 &  5.9 & 11.1 \\
\modelicon{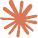}Claude Sonnet 4.6      & 60.2 & 51.7 & 24.3 & 34.6 & 57.9 & 33.0 & 23.5 & 65.6 & 37.6 & 16.1 & 60.3 & 32.6 & 22.2 & 75.9 & 20.4 & 11.1 & 71.1 & 16.1 & 12.3 \\
\modelicon{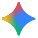}Gemini 2.5 Flash       & 98.7 & 47.1 & 21.8 & 32.2 & 94.4 & 39.0 & 23.0 & 97.2 & 36.0 & 13.8 & 92.6 & 31.9 & 18.4 & 91.4 & 15.8 & 10.3 & 92.0 &  9.6 & 10.3 \\
\modelicon{figures/openai.pdf}GPT-5.4 Mini           & 94.1 & 43.8 &  5.8 & 11.5 & 89.1 & 36.2 &  5.8 & 96.0 & 35.8 &  9.2 & 93.7 & 28.1 &  4.6 & 82.6 & 16.5 &  1.1 & 86.1 &  7.7 &  0.0 \\
\midrule
\multicolumn{20}{@{}l}{\textit{Open-source}} \\
\modelicon{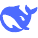}DeepSeek V4 Pro        & 59.9 & 38.5 & \textbf{37.0} & 49.4 & 59.2 & 39.1 & \textbf{38.3} & 64.2 & 42.5 & \textbf{37.0} & 59.0 & 38.5 & \textbf{34.6} & 75.5 & 27.8 & \textbf{22.2} & 50.9 & 25.3 & 19.8 \\
\modelicon{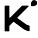}Kimi K2.5              & 84.4 & 44.1 & 24.7 & 37.0 & 68.7 & 28.7 & 21.0 & 79.7 & 38.7 & 29.6 & 74.2 & 36.2 & 25.9 & 84.7 & 17.0 & 12.3 & 83.3 & 14.2 & 12.3 \\
\modelicon{figures/qwen-color.pdf}Qwen3-14B (base)       & 98.2 & 12.4 &  4.9 &  9.9 & 93.2 & 12.4 &  4.9 & 98.3 & 13.1 &  6.2 & 97.9 & 10.9 &  1.2 &100.0 &  0.4 &  0.0 &100.0 &  0.2 &  0.0 \\
\modelicon{figures/qwen-color.pdf}Qwen3-235B-A22B        &100.0 & 23.1 &  3.7 &  6.2 & 96.8 & 21.5 &  3.7 & 98.9 & 20.8 &  4.9 & 90.4 & 17.0 &  3.7 & 62.5 & 11.5 &  3.7 & 63.9 & 10.6 &  0.0 \\
\midrule
\multicolumn{20}{@{}l}{\textit{Domain-specific}} \\
LlaSMol-7B             & 57.1 &  5.2 &  1.2 &  1.2 & 81.5 & 27.8 &  0.0 & 83.1 & 30.5 &  0.0 &  0.0 &  0.0 &  0.0 &  0.0 &  0.0 &  0.0 &  0.0 &  0.0 &  0.0 \\
ChemDFM-v2.0-14B       & 93.9 & 49.5 &  0.0 &  0.0 & 85.1 & 46.6 &  0.0 & 96.6 & 52.5 &  0.0 & 91.6 & 39.4 &  1.2 & 59.6 & 32.8 &  0.0 & 67.7 & 25.6 &  0.0 \\
ChemLLM-20B            & 89.0 & 54.1 &  0.0 &  0.0 & 86.3 & 41.2 &  0.0 & 88.8 & 51.8 &  0.0 & 87.5 & 58.4 &  0.0 & 87.2 & 30.8 &  0.0 & 79.5 & 30.8 &  0.0 \\
\bottomrule
\end{tabular}%
}
\vspace{-0.15in}
\end{table*}

\section{Analysis and Findings}
\subsection{When Does Tool Use Become Effective Procurement Reasoning?}
\label{5.1}
\vspace{-0.1in}
\begin{figure*}[h]
\centering
\includegraphics[width=\linewidth]{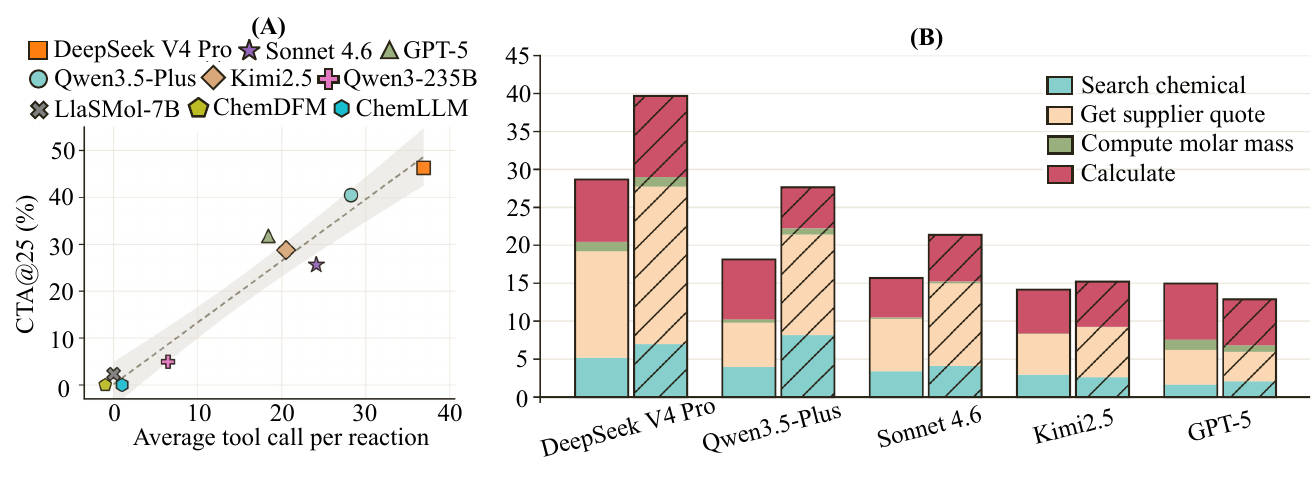}
\vspace{-0.35in}
\caption{\textbf{Tool calling analysis.} \textbf{(A)} Average tool calls versus CTA@25 across ReAct agents, with an Ordinary Least Squares fit and 95\% confidence band. \textbf{(B)} Tool-call composition for the strongest agents, where solid bars denote successful trajectories and hatched bars denote failed trajectories.}
\label{fig:tool-usage}
\end{figure*}

Figure~\ref{fig:tool-usage} analyzes how ReAct agents interact with the ChemCost tool environment. The results separate agents into three tool-use regimes, reflecting limited engagement with the tool interface, frequent tool use without reliable cost recovery, and more complete execution of the procurement workflow. Successful agents must determine which chemicals to ground, retrieve the relevant supplier quotes, select packs that satisfy the constraints, and aggregate the evidence into a final cost estimate.

\textbf{Tool engagement is necessary but not sufficient.} Figure~\ref{fig:tool-usage} (A) shows a strong across-model association between mean tool calls per reaction and CTA@25, with Pearson correlation \(r=0.98\) across ReAct models. This correlation mainly reflects whether a model enters the tool-use workflow at all. Models near zero tool calls, including chemistry-specialized models and Qwen3-235B, remain near zero CTA@25, whereas active frontier agents achieve substantially higher accuracy. After removing the near-zero-call cluster, the correlation remains positive but weaker, around \(r=0.77\). Thus, tool interaction is a prerequisite for meaningful ChemCost performance, but aggregate call count alone does not explain successful procurement reasoning.

\textbf{More tool calls do not necessarily imply better trajectories.}
Figure~\ref{fig:tool-usage} (B) shows that, for most frontier models, failed trajectories use more tools than successful trajectories. Failed cases require \(38\%\) more calls for DeepSeek V4 Pro, \(52\%\) more for Qwen3.5-Plus, \(36\%\) more for Sonnet 4.6, and \(7\%\) more for Kimi K2.5. These excess calls are concentrated largely in supplier-quote retrieval, suggesting that many failures are not due to too few retrieval attempts. Instead, agents often continue querying without converging to a valid pack choice or a complete procurement computation. Tool-call volume therefore captures engagement, but not necessarily effective reasoning.

\textbf{Tool-use failures have distinct mechanisms.}
The same final error can arise from different tool-use patterns. Frontier models often exhibit a stuck-retrieval pattern, repeatedly querying supplier quotes without resolving usable procurement evidence. GPT-5 shows a contrasting insufficient-retrieval pattern, with failed trajectories using approximately \(14\%\) fewer calls than successful ones. This suggests earlier termination or reliance on parametric estimates before enough evidence is recovered. These patterns indicate that ChemCost separates several axes of scientific tool use, including workflow ordering, retrieval persistence, grounding quality, and post-retrieval reasoning.

\subsection{How Does Input Noise Change Agent Failure Modes?}
\vspace{-0.2in}
\begin{figure*}[h]
\centering
\includegraphics[width=\linewidth]{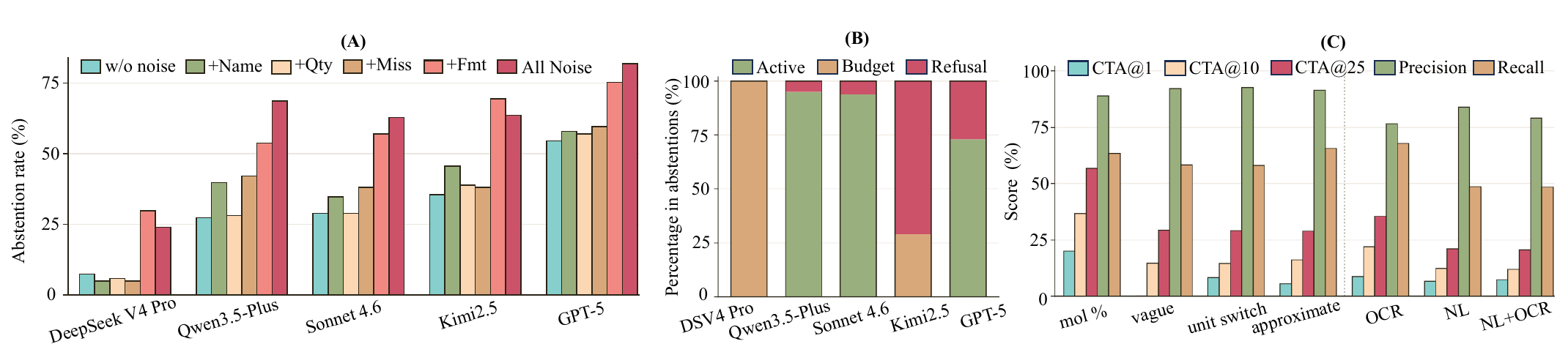}
\vspace{-0.3in}
\caption{\textbf{Noise effect on ReAct agents.} \textbf{(A)} Abstention rates for agents under the clean and noise-injected views. \textbf{(B)} Decomposition of \emph{+Fmt} abstentions into active non-completion, budget exhaustion, and explicit refusal.
\textbf{(C)} Fine-grained analysis of quantity and format perturbations.}
\label{fig:noise}
\end{figure*}

Figure~\ref{fig:noise} evaluates the five frontier ReAct agents under five controlled noise conditions: chemical-name perturbation (\emph{+Name}), quantity re-expression (\emph{+Qty}), missing-field perturbation (\emph{+Miss}), format perturbation (\emph{+Fmt}), and their joint application (\emph{All Noise}). Ground-truth procurement labels are held fixed across conditions, so changes in performance reflect robustness to input perturbation rather than changes in item difficulty.

\textbf{Format noise is the dominant abstention trigger.}
Figure~\ref{fig:noise}A shows that \emph{+Fmt} produces the largest increase in abstention across all five frontier agents. Abstention rises from \(7.4\%\) to \(29.8\%\) for DeepSeek V4 Pro, from \(27.3\%\) to \(53.7\%\) for Qwen3.5-Plus, from \(28.9\%\) to \(57.0\%\) for Sonnet~4.6, from \(35.5\%\) to \(69.4\%\) for Kimi K2.5, and from \(54.5\%\) to \(75.2\%\) for GPT-5. On average, \emph{+Fmt} increases abstention by \(26.3\) percentage points, while no other individual noise stage exceeds \(6\) percentage points. This indicates that realistic presentation changes, such as prose formatting and OCR-like corruption, primarily disrupt the earliest part of the workflow, where the agent must parse chemicals, quantities, roles, and product information before any procurement reasoning can begin.

\textbf{The same abstention outcome reflects different trajectory mechanisms.}
Figure~\ref{fig:noise}B decomposes \emph{+Fmt} abstentions into active non-completion, budget exhaustion, and explicit refusal. Although all three mechanisms produce the same outward result, namely no valid scalar prediction, their distributions differ sharply across models. DeepSeek V4 Pro's abstentions are dominated by budget exhaustion, suggesting repeated recovery attempts that fail to converge within the step limit. Qwen3.5-Plus and Sonnet~4.6 are dominated by active non-completion, with approximately \(95\%\) of abstentions in this category. Kimi K2.5 is mostly refusal-driven, with \(71\%\) refusal and \(29\%\) budget exhaustion, while GPT-5 mixes active non-completion and refusal. Thus, format noise does not induce a single generic failure mode. It exposes model-specific regimes of failed search, incomplete task engagement, and explicit refusal.

\textbf{Noise opens a recognition-computation gap.}
Figure~\ref{fig:noise}C compares CTA@25 with component precision and recall under fine-grained \emph{+Qty} and \emph{+Fmt} perturbations. Component precision remains high, between \(77\%\) and \(93\%\), even when CTA@25 falls to \(21\%\). The gap between recognizing chemicals and computing procurement cost widens, with the CTA@25-to-precision ratio decreasing from \(0.64\) under mol\% perturbation to \(0.26\) under NL+OCR. Recall decreases, from \(63\%\) under mol\% to \(49\%\) under NL+OCR, but the CTA@25 drop is larger. These results show that noisy-input failures are not merely recognition failures. Agents often identify plausible components, but fail to assemble them into a complete and correctly normalized procurement state.

\subsection{Which Reaction Structures Make Procurement Reasoning Difficult?}
\vspace{-0.2in}
\begin{figure*}[h]
\centering
\includegraphics[width=\linewidth]{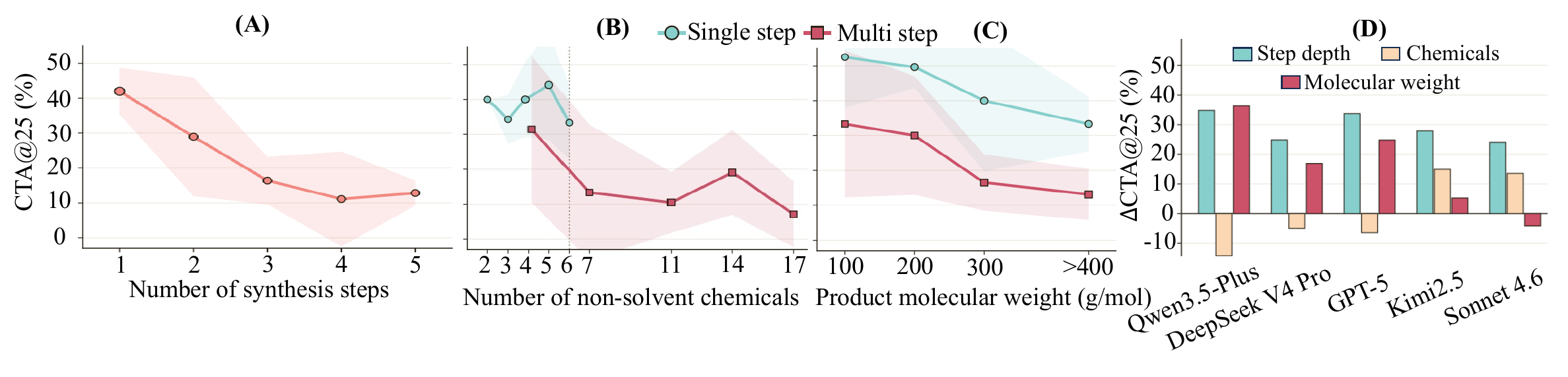}
\vspace{-0.3in}
\caption{\textbf{Reaction-type difficulty along four observable axes.}
\textbf{(A)} CTA@25 as a function of synthesis depth, with shaded 95\% bootstrap confidence intervals.
\textbf{(B)} CTA@25 versus the number of chemicals.
\textbf{(C)} CTA@25 versus product molecular weight.
\textbf{(D)} Per-model \(\Delta\)CTA@25, defined as CTA@25(easier) \(-\) CTA@25(harder). \textit{Step depth} (single-step vs.\ multi-step \(\geq 3\)), \textit{Chemicals} (single-step reactions with \(\leq 3\) vs.\ \(\geq 5\) non-solvent components), and \textit{Molecular weight} (single-step reactions with low vs.\ high product MW).}
\label{fig:reaction-type}
\vspace{-0.1in}
\end{figure*}


\textbf{Route depth is the dominant structural difficulty axis.}
Figure~\ref{fig:reaction-type}(A--C) shows that ChemCost difficulty is structured along more than one axis. Route depth is the strongest factor. CTA@25 drops from \(42\%\) for single-step reactions to \(29\%\), \(16\%\), \(11\%\), and \(13\%\) for two- through five-step routes. By contrast, component burden alone is not a strong independent signal within single-step reactions, where CTA@25 remains relatively flat across two to six non-solvent components. Product molecular weight adds a separate gradient, with accuracy decreasing for both single-step and multi-step reactions as molecular weight increases. The consistent gap between single-step and multi-step reactions within each molecular-weight bin suggests that route depth and product scale contribute partially independent difficulty signals.

\textbf{Models have distinct reaction-complexity fingerprints.}
Figure~\ref{fig:reaction-type}(D) decomposes these effects by model. The depth effect is universal where every model loses substantial CTA@25 from low-depth to high-depth reactions, with drops of roughly 24 to 25 percentage points. By contrast, the single-step chemical-count and molecular-weight effects differ across models. Qwen3.5-Plus, DeepSeek V4 Pro, and GPT-5 do not degrade with more chemicals in single-step reactions; in these models, higher component count is associated with equal or better accuracy. One possible explanation is that multi-component single-step records often provide a more complete reaction recipe, including common catalysts, ligands, bases, or additives that help anchor retrieval and pack selection. Kimi K2.5 and Sonnet 4.6 show the more intuitive pattern, where higher single-step component count reduces accuracy, suggesting that their bottleneck is more sensitive to component ordering, retrieval coverage, or pack selection.

\vspace{-0.1in}
\section{Conclusion and Future Work}
We introduced \textsc{ChemCost}, a benchmark for evaluating scientific tool-use agents on chemical procurement cost estimation with deterministic labels. The task requires agents to ground chemicals, retrieve frozen supplier quotes, select valid packs, normalize quantities, and compute costs from reaction descriptions. Experiments show that tool access is necessary but insufficient, since failures often arise from poor evidence integration and realistic format noise substantially increases abstention. Future work can extend \textsc{ChemCost} with richer procurement constraints, including stock availability, shipping, regional pricing, lead time, and alternative pack-selection rules. Progress on agents will require stronger chemical grounding, retrieval-aware state tracking, calibrated abstention, and reflection over intermediate tool evidence.

\section*{Acknowledgments}

This work was supported by the U.S. National Science Foundation (NSF) under Award \#2607594 (PCLNet: The AI Science Foundry at Carnegie Mellon University). O.I. is supported by the National Science Foundation through the Center for Computer-Assisted Synthesis (C-CAS) under Award CHE-2202693. This work used the Bridges-2 (PSC) and Delta (NCSA) systems through allocations from the Advanced Cyberinfrastructure Coordination Ecosystem: Services \& Support (ACCESS) program, which is supported by NSF grants \#2138259, \#2138286, \#2138307, \#2137603, and \#2138296.

\bibliography{references}
\bibliographystyle{plain}

\medskip


\appendix
\clearpage
\appendix
\section{Limitations}
\label{limitations}
\textsc{ChemCost} is a controlled benchmark for scientific tool use and intentionally abstracts from several aspects of real-world procurement. Its labels measure a fixed \(1\) mmol limiting-reagent scale normalized by product mass, using a frozen supplier snapshot, a \(95\%\) purity threshold, and a fixed pack-selection rule. This design makes evaluation deterministic and comparable across reactions, but does not model stock availability, shipping, regional pricing, lead time, bulk discounts, or alternative purchasing preferences. The current experiments also focus on ReAct-style agents with a fixed tool set and step budget, so future work can extend the benchmark with richer procurement constraints, broader supplier coverage, stronger agent controllers, and larger human calibration studies.

\section{Source-specific extraction and evaluability}
We summarizes how reactions are extracted from each source family and which sources contribute to the evaluable set in ~\ref{tab:sources}.

\begin{table*}[h]
\centering
\small
\caption{Source-specific extraction and evaluability rules. Yield is
required for procurement-cost ground truth (denominator
$\mathrm{grams\_product} = \mathrm{MW}_p \times 0.001 \times \mathrm{yield}/100$).}
\label{tab:sources}
\setlength{\tabcolsep}{6pt}
\begin{tabular}{@{}l l l c@{}}
\toprule
Source & Extraction rule & Yield source & Evaluable \\
\midrule
ORD               & Schema + unit conversion   & Reported          & \checkmark \\
Textbooks         & Regex on reagent blocks    & Extracted         & \checkmark \\
Organic Syntheses & Regex + step markers       & Extracted         & \checkmark \\
ChemPU            & XDL action tags            & Extracted if any  & \checkmark \\
PaRoutes          & Route-tree traversal       & ---               & \checkmark \\
Hand-crafted      & Authored templates         & Specified         & \checkmark \\
\bottomrule
\end{tabular}
\end{table*}

\section{Stage-level Error Decomposition}
\label{appendix:stage-pipeline}

We decompose each ReAct trajectory on the Clean set into four conditional stages that mirror the \textsc{ChemCost} procurement workflow.
\begin{itemize}[leftmargin=*,itemsep=2pt]
\item \textbf{S1 Grounding}, where a ground-truth component is matched in \texttt{predicted\_components} through an alias-aware canonical-name match.
\item \textbf{S2 Retrieval $\mid$ S1}, where the agent issues a \texttt{get\_supplier\_quotes} call whose argument resolves to the same database \texttt{chemical\_id} as the matched ground-truth component.
\item \textbf{S3 Pack Selection $\mid$ S2}, where the agent selects a procurement option consistent with the oracle pack-selection rule, within a 5\% tolerance in component procurement cost.
\item \textbf{S4 Aggregation}, where the agent's reported \texttt{predicted\_cost\_per\_gram} is reproduced within 10\% from its predicted component prices using the benchmark mass and yield arithmetic.
\end{itemize}

We do not isolate per-component mass normalization as a separate stage because agents do not reliably output per-component masses. Errors in mass normalization are therefore reflected indirectly through pack-selection and aggregation failures, but cannot be uniquely localized from the logged trajectories alone.

\begin{table}[h]
\centering
\small
\caption{Stage-level error decomposition for five frontier ReAct agents on the Clean diagnostic set. Each cell reports the conditional pass-through rate of a stage given successful upstream stages, except S4, which is computed at the reaction level. Conditional rates should be interpreted together with upstream coverage. Best per column in \textbf{bold}.}
\label{tab:stage-pipeline}
\setlength{\tabcolsep}{8pt}
\begin{tabular}{@{}l cccc@{}}
\toprule
Model & S1 Grounded & S2\,$|$\,S1 Retrieved & S3\,$|$\,S2 Pack & S4 Aggregation \\
\midrule
Qwen3.5-Plus      & 49.6 & 76.3 & 18.4 & \textbf{33.8} \\
GPT-5             & 15.6 & 67.0 &  1.8 & 12.1 \\
Claude Sonnet 4.6 & 47.2 & 72.9 & 58.6 &  3.1 \\
DeepSeek V4 Pro   & \textbf{85.1} & 54.3 & \textbf{66.2} & 21.6 \\
Kimi K2.5         & 35.0 & \textbf{78.7} & 48.3 & 16.9 \\
\bottomrule
\end{tabular}
\end{table}

The five frontier agents fail at different stages of the procurement pipeline. DeepSeek V4 Pro grounds most components, but a large fraction of grounded components are not followed by quote retrieval. Qwen3.5-Plus retrieves quotes reliably once components are grounded, but frequently fails the pack-selection rule. Claude Sonnet 4.6 reaches reasonable component-level pack selection, yet rarely produces an aggregation consistent with the benchmark arithmetic. GPT-5 has the weakest first-stage coverage, reflecting its higher abstention and early termination rate. These heterogeneous failure profiles support stage-targeted improvements rather than a single generic fix for scientific tool use.



\section{Human Reference Protocol}
\label{appendix:human}

Table~\ref{tab:main} reports a Human reference row alongside the ReAct agents. We document here how the row was produced and how it should be interpreted.

\paragraph{Goal of the human reference.}
The Human row is intended as a \emph{database-grounded expert reference} on \textsc{ChemCost}, not as an estimate of unaided human chemistry knowledge or a theoretical upper bound. Concretely, we ask whether trained chemists can recover the deterministic procurement cost under the same agent-facing input view, frozen supplier database, fixed pack-selection rule, and purity threshold used by the ReAct agents. This frames the Human row as an expert reference under the benchmark information state and procurement rules.

\paragraph{Participants.}
The Human reference was produced by three trained chemists with prior wet-lab synthesis experience, recruited internally. Each participant held a graduate-level degree in chemistry or chemical engineering and reported routine familiarity with commercial supplier catalogues. Participation was uncompensated and treated as authorial contribution; no external crowd workers were involved. No personal or sensitive participant data were collected.

\paragraph{Inputs available.}
Each participant received the same agent-facing reaction view used by the ReAct agents, including component names, component roles, stoichiometric equivalents, molecular weights, reported yield, and product identity. Supplier prices and canonical molecular identifiers, such as SMILES and CAS numbers, were withheld from the input. Participants were permitted to consult public chemical references for name resolution and to query the same frozen \textsc{ChemCost} supplier database used to compute ground truth. They applied the same purity threshold and fixed pack-selection rule. Participants were not permitted to consult an LLM during estimation.

\paragraph{Procedure.}
Each evaluated reaction was assigned to exactly one of the three participants under round-robin assignment, with no overlap. The same assignment was used across noise conditions so that noise effects are computed within participant. Participants returned a single procurement-cost number per reaction together with the component identities and pack selections supporting the estimate. Abstentions are counted as failures under CTA-based scoring and are reported separately when applicable. There was no per-reaction time limit; total annotation time for the reported reference was approximately 30 hours.

\paragraph{Scoring.}
Human predictions were scored using the same CTA@$k$ metrics as the automated agents. The reported row in Table~\ref{tab:main} is the per-reaction CTA@$k$ rate over all evaluated reactions, including abstentions as failures. Because each reaction was assigned to a single participant, no inter-annotator averaging is performed. The Human row is provided as an expert reference and is not included in per-column boldface comparisons among automated agents.

\section{Reasoning Method Analysis}
\label{sec:reasoning-methods}
\begin{table*}[h]
\centering
\small
\caption{Reasoning method analysis on the all set. Same-backbone comparison: each block reports ZeroShot, FewShot, Chain-of-Thought (no tools), and ReAct (with tools) on the same LLM. The \textbf{Tools} column indicates whether the agent invokes the four-tool ChemCost stack. Cells report component Precision, Recall, and CTA at three relative-error thresholds (\textbf{C.1} = strict tolerance, \textbf{C.10}, \textbf{C.25}). Best per backbone block in \textbf{bold}.}
\label{tab:reasoning-methods}
\setlength{\tabcolsep}{8pt}
\begin{tabular}{@{}l l c cc ccc@{}}
\toprule
Backbone        & Method            & Tools          & Precision      & Recall         & C.1            & C.10           & C.25 \\
\midrule
\multicolumn{8}{@{}l}{\textit{Sonnet 4.6}} \\
                & ZeroShot          & $\times$       & 48.4           & 8.3            & 0.0            & 0.0            & 0.8 \\
                & FewShot           & $\times$       & 74.0           & 23.8           & 0.0            & 3.3            & 4.1 \\
                & Chain-of-Thought  & $\times$       & \textbf{79.5}  & \textbf{52.9}  & 0.8            & 1.7            & 2.5 \\
                & ReAct             & $\checkmark$   & 60.2           & 51.7           & \textbf{11.1}  & \textbf{24.7}  & \textbf{34.6} \\
\midrule
\multicolumn{8}{@{}l}{\textit{DeepSeek V4 Pro}} \\
                & FewShot           & $\times$       & 79.7           & 77.8           & 0.8            & 0.8            & 2.5 \\
                & Chain-of-Thought  & $\times$       & \textbf{81.8}  & \textbf{87.2}  & 0.0            & 0.8            & 4.1 \\
                & ReAct             & $\checkmark$   & 59.9           & 38.5           & \textbf{14.8}  & \textbf{37.0}  & \textbf{49.4} \\
\midrule
\multicolumn{8}{@{}l}{\textit{Qwen3.5-Plus}} \\
                & ZeroShot          & $\times$       & 70.2           & \textbf{98.4}  & 0.0            & 0.8            & 1.7 \\
                & FewShot           & $\times$       & 81.8           & 88.3           & 0.0            & 4.1            & 7.4 \\
                & Chain-of-Thought  & $\times$       & 82.8           & 90.8           & 0.8            & 3.3            & 4.1 \\
                & ReAct             & $\checkmark$   & \textbf{100.0} & 54.3           & \textbf{16.0}  & \textbf{34.6}  & \textbf{50.6} \\
\bottomrule
\end{tabular}
\end{table*}

\paragraph{Experimental setup.}
We isolate two design variables: tool availability and reasoning structure. On three frontier backbones, Sonnet~4.6, DeepSeek V4 Pro, and Qwen3.5-Plus, we compare four methods on the all set: \emph{ZeroShot}, which directly predicts the cost without tools; \emph{FewShot}, which prepends worked procurement examples without tools; \emph{Chain-of-Thought} (CoT), which prompts step-by-step reasoning without tools; and \emph{ReAct}, which uses the full ChemCost tool stack for chemical search, supplier-quote retrieval, molecular-weight computation, and arithmetic. All methods receive the same reaction input and are evaluated against the same deterministic labels. This setup asks whether procurement accuracy can be recovered by reasoning prompts alone, or whether external grounding through tools is necessary.

\textbf{Tool access is necessary for cost accuracy.}
ReAct outperforms every tool-free method by a large margin on the headline metric. Sonnet~4.6 reaches \(34.6\%\) CTA@25 with ReAct, compared with a best tool-free score of \(4.1\%\). DeepSeek V4 Pro reaches \(49.4\%\) with ReAct, compared with \(4.1\%\) under CoT, and Qwen3.5-Plus reaches \(50.6\%\), compared with \(7.4\%\) under FewShot. The gap is even more pronounced at the strict CTA@1 threshold: no tool-free method exceeds \(0.8\%\) on any backbone, while ReAct attains \(11\%\)--\(16\%\). These results indicate that parametric knowledge and prompt-only reasoning are insufficient for gram-scale procurement-cost prediction. Accurate performance requires external retrieval of chemical identities and supplier prices.

\textbf{Reasoning prompts improve recognition more than computation.}
Tool-free reasoning methods can recover components, but this does not translate into accurate costs. For example, CoT substantially increases component recall for Sonnet~4.6, from \(8.3\%\) under ZeroShot to \(52.9\%\), yet CTA@25 remains below \(3\%\). Qwen3.5-Plus provides an even clearer separation: its ZeroShot run reaches \(98.4\%\) component recall but only \(1.7\%\) CTA@25. Across all three backbones, CTA@25 changes by at most a few percentage points among ZeroShot, FewShot, and CoT. This shows that the main bottleneck is not simply recognizing which chemicals appear in the reaction. Without supplier retrieval, pack selection, and cost aggregation, better textual reasoning does not produce reliable procurement estimates.

\textbf{ReAct trades broad recall for retrieval-verified groundedness.}
ReAct often reports fewer components than CoT, but the components it reports are more tightly grounded. Component recall drops from CoT to ReAct on DeepSeek V4 Pro (\(87.2\%\to 38.5\%\)) and Qwen3.5-Plus (\(90.8\%\to 54.3\%\)), while remaining roughly stable on Sonnet~4.6 (\(52.9\%\to 51.7\%\)). This drop should not be interpreted as a loss of chemical knowledge. For Qwen3.5-Plus, ReAct reaches \(100.0\%\) component precision, meaning that every emitted component is ground-truth correct. Rather, ReAct only produces a usable component when the model successfully maps a surface chemical name to a canonical entry and uses it in the procurement workflow. The limiting factor is therefore retrieval-grounded computation, not chemical recognition alone.

\section{Computing Resources}
\label{appendix:compute}

All dataset construction, evaluation, and local model inference were run on an internal compute node equipped with 8 NVIDIA H100 GPUs. ChemCost does not require model training; the compute cost comes primarily from agent inference, tool execution, trajectory logging, and post-hoc evaluation. Dataset preprocessing, chemical grounding validation, supplier-database indexing, ground-truth cost computation, and metric aggregation were run on CPU, with GPU resources used only for local LLM inference.

For proprietary or hosted models, including GPT-5, Claude Sonnet 4.6, Gemini 2.5 Flash, Qwen3.5-Plus, and GPT-5.4 Mini, model inference was performed through the corresponding API providers. Our local compute for these models was limited to running the ReAct controller, executing deterministic tools, storing tool trajectories, and computing metrics. For open-weight and chemistry-specialized models, including Qwen3-14B, Qwen3-235B-A22B, LlaSMol-7B, ChemDFM-v2.0-14B, and ChemLLM-20B, inference was run locally on the 8$\times$H100 node. Smaller models were evaluated on one or a few H100 GPUs, while larger models used multi-GPU inference with tensor parallelism when required.

Each ReAct run used a maximum budget of 40 reasoning-action steps per reaction. The dominant computational cost is therefore proportional to the number of evaluated reactions, the number of noise views, and the number of tool calls made by each model. Since all tools are deterministic and lightweight relative to model inference, GPU usage is driven mainly by LLM generation rather than supplier lookup, molecular-weight computation, or arithmetic evaluation. No gradient-based training, fine-tuning, or reinforcement learning was performed for the experiments in this paper.

\section{Agent Prompts}
\label{appendix:prompts}

This section documents the prompts used for each agent class in
Table~\ref{tab:main} and Table~\ref{tab:reasoning-methods}. All prompt
classes receive the same agent-facing reaction view for a given evaluation
condition. In the clean condition, this view contains component names, roles,
equivalents, molecular weights, reported yield, and product identity. In
noise-injected conditions, fields are perturbed or omitted according to the
noise protocol. Supplier prices and canonical molecular identifiers are
withheld from every prompt class.

\paragraph{Component-list rendering.}
For every prompt class, the structured component list embedded in the user
prompt is rendered as one row per component using:
\begin{quote}\small\ttfamily
- \{name\} | role: \{role\} | equiv: \{equivalents\} | MW: \{mw\} g/mol
\end{quote}
When a component has a free-text quantity description, as in the \emph{+Qty}
setting, \texttt{equiv: \{value\}} is replaced by
\texttt{qty: \{quantity\_description\}}. Solvents are included in the rendered
list and explicitly marked as excluded from cost.

\paragraph{ReAct system prompt.}
The ReAct agent receives a system prompt that specifies the cost model, the
four tools, the pack-selection rule, and the output format. The full text is
reproduced below:

\begin{quote}\footnotesize\ttfamily
Your task is to estimate the fixed-scale procurement cost normalized per gram of product for a given reaction. Use a 1 mmol limiting-reagent reaction scale.

\medskip\noindent\#\# Cost Model

The benchmark uses a procurement cost model: how much does it cost to
purchase the required non-solvent components for this reaction at laboratory
scale?

\medskip\noindent Fixed rules:\\
- Reaction scale: 1 mmol of the limiting reagent.\\
- Required mass per component: equivalents $\times$ MW $\times$ 0.001 grams.\\
- Purchase cost: select the smallest commercially available pack that covers
the required mass, using only packs with purity $\ge$ 95\%.\\
- If the required mass exceeds every available pack, buy the minimum number
of largest packs needed to cover the required mass.\\
- Solvents with role = "solvent" are excluded from cost.\\
- Yield affects only the denominator, namely grams of product produced.\\
- Catalytic mol\% must be converted as equiv = mol\% / 100.

\medskip\noindent Formula:\\
required\_mass\_g\_i = equiv\_i $\times$ MW\_i $\times$ 0.001\\
purchase\_cost\_usd\_i = selected pack price for component i\\
total\_purchase\_usd = sum of purchase\_cost\_usd\_i over non-solvent components\\
grams\_of\_product = product\_MW $\times$ 0.001 $\times$ (yield / 100)\\
procurement\_cost\_per\_g = total\_purchase\_usd / grams\_of\_product

\medskip\noindent\#\# Tools available\\
- \texttt{search\_chemical(query)}: resolve a chemical name to candidate
chemical identities and molecular weights.\\
- \texttt{get\_supplier\_quotes(smiles\_or\_name)}: retrieve frozen supplier
pack quotes with quantity\_g, price\_usd, and purity.\\
- \texttt{compute\_molar\_mass(smiles)}: compute molecular weight from SMILES
using RDKit.\\
- \texttt{calculate(expression)}: perform arithmetic.

\medskip\noindent\#\# Pack selection rules\\
1. Use only quotes with purity $\ge$ 95\%.\\
2. Select the smallest pack with quantity\_g $\ge$ required\_mass\_g.\\
3. If no single pack covers the required mass, buy
$\lceil$required\_mass\_g / largest\_pack\_g$\rceil$ units of the largest pack.\\
4. The selected pack price, or the total price of repeated largest packs, is
the component purchase cost.

\medskip\noindent This is a non-interactive benchmark. Do not ask clarifying
questions. Return a JSON answer. If you cannot complete the estimate, return
\verb|{"predicted_cost_per_gram": null, "predicted_components": []}|.

\medskip\noindent Respond with the final answer in JSON:\\
\verb|{"predicted_cost_per_gram": <number or null>,|\\
\verb| "predicted_components": [|\\
\verb|   {"name": <str>,|\\
\verb|    "selected_pack_g": <number or null>,|\\
\verb|    "selected_pack_price_usd": <number or null>}|\\
\verb| ]}|
\end{quote}

The ReAct user message is rendered from the following template:
\begin{quote}\footnotesize\ttfamily
Use a 1 mmol limiting-reagent scale and report the resulting procurement cost divided by the grams of product obtained at that scale.

\medskip\noindent Reaction: \{reaction\_name\}\\
Product MW: \{product\_mw\} g/mol\\
Yield: \{yield\_percent\}\%

\medskip\noindent Components. Solvents should be excluded from cost:\\
\{components\_text\}

\medskip\noindent Use the available tools to resolve chemical identities,
retrieve supplier pack options, select valid packs, and compute the final
procurement cost per gram.
\end{quote}
The ReAct loop runs for up to 40 model-tool turns per reaction.

\paragraph{ZeroShot prompt.}
The ZeroShot baseline receives the same reaction view as ReAct, but all
tool-use instructions are removed. The model receives no tools and no worked
examples. The prompt asks the model to estimate the cost in USD from
the provided reaction information and its parametric knowledge, then return
the same JSON schema.

\paragraph{Chain-of-Thought prompt.}
The Chain-of-Thought baseline receives the same tool-free reaction view as
ZeroShot, together with an explicit calculation procedure. The procedure asks
the model to identify non-solvent components, convert mol\% to equivalents
when needed, estimate component purchase costs, compute required masses at
1 mmol scale, sum component costs, compute grams of product from product MW
and yield, and divide. The model is asked to provide a concise calculation
trace followed by the final answer as valid JSON. No tools are available.

\paragraph{Few-Shot prompt.}
The Few-Shot baseline reuses the tool-free Chain-of-Thought prompt and
prepends two worked procurement-cost examples before the target reaction. The
examples illustrate limiting-reagent identification, required-mass computation
at 1 mmol scale, small-pack purchase-cost estimation, total purchase-cost
aggregation, product-mass computation, and final cost-per-gram calculation.
No tools are available.

\paragraph{Few-Shot ReAct prompt.}
The Few-Shot ReAct baseline combines the ReAct system prompt and tools with
one worked tool-use example. The example demonstrates chemical name
resolution, supplier-quote retrieval, fixed-rule pack selection, and final
cost aggregation. The target user prompt is identical to the ReAct user prompt.

\paragraph{Provider- and model-specific decoding.}
Across providers, we keep the task prompt, tool descriptions, cost model, and
JSON output schema fixed. Provider-specific differences are limited to API
parameters required to execute the same ReAct loop. For OpenAI-compatible
chat-completion models, we use temperature~0 when supported. For reasoning
models or hosted endpoints that do not expose temperature control, we use the
lowest-variance setting supported by the provider. We allocate 8K--16K
completion tokens depending on the provider limit and use the same 40-step
ReAct budget for all tool-augmented agents. No model receives additional
task-specific hints beyond the shared prompt family described above.

\section{Dataset Statistics and Filtering Flow}
\label{appendix:dataset-flow}

\textsc{ChemCost} is curated through a sequence of source-specific extraction,
chemical grounding, supplier-coverage, and evaluator-readiness checks. These
filters convert raw records from each source family into the 1{,}427 evaluable
reactions used in the main experiments. Table~\ref{tab:dataset-flow} reports
the per-source filtering flow. Route-level sources are reported at the route
level, while single-step sources are reported at the reaction level.

\paragraph{Filter definitions.}
The columns of Table~\ref{tab:dataset-flow} are computed as follows.
\begin{itemize}[leftmargin=*,itemsep=2pt]
\item \textbf{Raw} denotes records harvested from each source before quality
filtering, including duplicates, malformed entries, and records with incomplete
component lists.
\item \textbf{Extracted} denotes records that pass the source-specific extractor
(Table~\ref{tab:sources}) and are converted into the unified \textsc{ChemCost}
reaction schema with a product entry and at least one non-solvent reactant.
\item \textbf{Yield-defined} denotes records with a yield value available for
oracle cost computation. For experimental literature sources, this is a parsed
measured yield. For route-library records, this may be a fixed nominal yield
used consistently in both the oracle label and the agent-facing input.
\item \textbf{MW-valid} denotes records whose product entry has a non-null
molecular weight after PubChem grounding and RDKit validation.
\item \textbf{Grounded} denotes records in which every non-solvent component
resolves to a canonical molecular identity through the PubChem-RDKit pipeline.
\item \textbf{Price-covered} denotes records in which every non-solvent
component has at least one valid pack-level supplier quote in the frozen pricing
database.
\item \textbf{Evaluable} denotes final records with a non-null
\texttt{procurement\_cost\_usd\_per\_g\_product}, at least one priced
non-solvent component, and internally consistent stoichiometry.
\end{itemize}

\begin{table}[h]
\centering
\small
\caption{Per-source filtering flow from raw records to the 1{,}427 evaluable
\textsc{ChemCost} reactions. Multi-step sources, including Organic Syntheses,
ChemPU, and PaRoutes, are aggregated at the route level; single-step sources
are reported at the reaction level. PaRoutes is shown separately because it uses
a fixed nominal yield rather than an experimentally measured yield. Hand-crafted
records are author-curated calibration entries included in the evaluable set.}
\label{tab:dataset-flow}
\setlength{\tabcolsep}{4.2pt}
\begin{tabular}{@{}l rrrrrrr@{}}
\toprule
Source & Raw & Extracted & Yield-defined & MW-valid & Grounded & Price-cov. & Evaluable \\
\midrule
ORD                  & 1{,}002 & 1{,}002 & 1{,}002 & 1{,}002 & 982 & 728 & 701 \\
Textbooks            &   817   &   817   &   817   &   662   & 654 & 612 & 588 \\
Organic Syntheses    &   123   &   123   &   123   &    44   &  29 &  14 &  11 \\
ChemPU               &    63   &    63   &    63   &    51   &  43 &  27 &  17 \\
PaRoutes             &   150   &   150   &   150   &   138   & 138 &  82 &  68 \\
Hand-crafted         &    50   &    50   &    50   &    50   &  48 &  42 &  42 \\
\midrule
\textbf{Total}       &\textbf{2{,}205}&\textbf{2{,}205}&\textbf{2{,}205}&\textbf{1{,}947}&\textbf{1{,}894}&\textbf{1{,}505}&\textbf{1{,}427}\\
\bottomrule
\end{tabular}
\end{table}

\paragraph{Source composition.}
The final benchmark combines four types of evidence. ORD and textbook records
provide broad single-step coverage with structured or semi-structured reaction
descriptions. Organic Syntheses and ChemPU contribute experimentally grounded
and route-level examples with more heterogeneous reporting conventions. PaRoutes
adds route-library examples that pass chemical grounding and supplier-coverage
checks and are evaluated using a fixed nominal yield consistently exposed to the
oracle and the agent. The hand-crafted subset contributes calibration reactions
designed to cover edge cases in stoichiometry, pack selection, and multi-step
cost aggregation.

\paragraph{Reconciling the headline count.}
The 1{,}427 evaluable reactions reported in the main text are the union of all
rows in the Evaluable column. Literature and database-derived sources contribute
1{,}317 evaluable records from ORD, textbooks, Organic Syntheses, and ChemPU.
The hand-crafted subset contributes 42 additional evaluable records, and the
PaRoutes route-library subset contributes 68 evaluable records with nominal
yield. All 1{,}427 records have deterministic oracle costs computed from the
same frozen supplier database and the same procurement rule.

\paragraph{Major filter losses.}
Three filters account for most record loss. First, yield parsing removes records
whose yield is absent, ambiguous, or not usable for per-gram cost normalization.
Second, molecular validation removes records whose product or component identity
cannot be resolved unambiguously through PubChem and RDKit. Third, supplier
coverage removes reactions in which at least one non-solvent component lacks a
valid pack-level quote in the frozen pricing snapshot. Records that fail price
coverage are excluded from the evaluable benchmark but can be retained in the
released metadata with a null procurement label, allowing future supplier
snapshots to recover additional instances without rerunning the full extraction
pipeline.


\newpage

\end{document}